\pdfoutput=1
\documentclass[10pt,twocolumn,letterpaper]{article}
\usepackage[utf8]{inputenc}
\usepackage{iccv}
\usepackage{times}
\usepackage{epsfig}
\usepackage{graphicx}
\usepackage{amsmath}
\usepackage{amssymb}

\usepackage{microtype}
\usepackage{subfigure}
\usepackage{booktabs} 
\usepackage{amsfonts}
\usepackage{multirow}
\usepackage{bm}
\def\x{\boldsymbol{x}}

\def\p{\boldsymbol{p}}
\def\q{\boldsymbol{q}}

\def\w{\boldsymbol{w}}

\DeclareMathOperator*{\argmax}{arg\,max}

\newtheorem{theorem}{Theorem}

\usepackage{marvosym}

\usepackage[pagebackref=true,breaklinks=true,letterpaper=true,colorlinks,bookmarks=false]{hyperref}

\iccvfinalcopy 

\def\iccvPaperID{6368} 

\ificcvfinal\fi

\begin{document}

\title{On Universal Black-Box Domain Adaptation}

\author{Bin Deng$^1$, Yabin Zhang$^2$, Hui Tang$^1$, Changxing Ding$^1$, Kui Jia\textsuperscript{\Letter}$^1$\\
$^1$South China University of Technology\\
$^2$Hong Kong Polytechnic University\\
}

\maketitle
\ificcvfinal\thispagestyle{empty}\fi

\begin{abstract}

In this paper, we study an arguably least restrictive setting of domain adaptation in a sense of practical deployment, where only the interface of source model is available to the target domain, and where the label-space relations between the two domains are allowed to be different and unknown. We term such a setting as Universal Black-Box Domain Adaptation (UB$^2$DA). The great promise that UB$^2$DA makes, however, brings significant learning challenges, since domain adaptation can only rely on the predictions of unlabeled target data in a partially overlapped label space, by accessing the interface of source model. To tackle the challenges, we first note that the learning task can be converted as two subtasks of in-class\footnote{In this paper we use in-class (out-class) to describe the classes observed (not observed) in the source black-box model.} discrimination and out-class detection, which can be respectively learned by model distillation and entropy separation. We propose to unify them into a self-training framework, regularized by consistency of predictions in local neighborhoods of target samples. Our framework is simple, robust, and easy to be optimized. Experiments on domain adaptation benchmarks show its efficacy. Notably, by accessing the interface of source model only, our framework outperforms existing methods of universal domain adaptation that make use of source data and/or source models, with a newly proposed (and arguably more reasonable) metric of H-score, and performs on par with them with the metric of averaged class accuracy. Code is available at \url{https://github.com/Gorilla-Lab-SCUT/UB2DA}.

\end{abstract}

\section{Introduction}
Unsupervised domain adaptation (UDA) aims to learn a prediction function for the unlabeled data of a target domain with the help of labeled data in a shifted but similar source domain. In the past decades, UDA has been heavily studied among different tasks such as object recognition \cite{Zhang_2020, zhang2020transfer}, semantic segmentation \cite{chen2019-da-segmentation, pan2020-da-segmentation}, object detection \cite{kin2019da-detection, xu2020-da-detection}, and person re-identification \cite{fu2019-da-reid, Ge2020Mutual}. Existing UDA approaches mainly focus on learning domain-invariant features \cite{long2015mmd, long2016unsupervised, ganin2016domain}, or taking the advantages of semi-supervised learning \cite{shu2018a, zhang2020label} and unsupervised learning \cite{tang2020-srdc} techniques. Apart from the close-set domain adaptation, where the same label space is shared across domains, problem variants of UDA have been introduced depending on the degree of overlap between the label space of the two domains. These options respectively give rise to other learning settings such as partial \cite{cao2018partial,zhang2018importance}, open-set \cite{saito2018open,busto2017openset}, and universal domain adaptations \cite{you2019universal, fu2020learning, saito2020dance}.
Among these UDA methods, source data are specifically required by target domain for the adaptation use.

Recently, source-free or source-absent domain adaptation methods \cite{liang2020shot, nath2020UniDA-sourcefree} are proposed due to increasing concerns for data privacy. They assume that the source data are not accessible to the target domain; a pre-trained source model is instead required and the adaptation is performed in a way similar to the hypothesis transfer learning (HTL) \cite{lao2020hypothesis}. In this paper, we consider a less restrictive learning setting, black-box domain adaptation, where only the interface of the source model is available for the target domain \cite{chidlovskii2016domain, zhang2021unsupervised}. Considering that increasing number of open-AI interfaces (e.g., GPT-3, Google and Tencent AI platforms etc.) are provided, it is convenient to directly use these off-the-shelf, black-box interfaces to assist the learning of our interested target models.
Such a learning strategy is really attractive since it not only preserves the data privacy but also maintains the commercial interests of providers of the black-box models.

However, when applying the black-box domain adaptation into practice, an inevitable challenge would be faced: it is hard to guarantee the task consistency between the target task and the fixed source one. In other words, the label space of the source black-box predictor may be not identical to the target one. The reason for this is two-fold: first, the target domain is fully unlabeled and thus we may know nothing about its label space; second, the target label set may be diverse in different applications whereas the source one is usually fixed. To this end, we propose a more practical setting of universal black-box domain adaptation (UB$^2$DA), where we further relax the restrictions by allowing the target label space to be varying and unknown. Different from the closed-set black-box domain adaptation \cite{chidlovskii2016domain,zhang2021unsupervised}, the proposed UB$^2$DA aims to learn a robust target prediction model that not only can recognize the in-class target examples but also can detect the out-class ones.


To address this challenge problem, we first note that the UB$^2$DA task can be converted as two subtasks of in-class discrimination and out-class detection. Specifically, the in-class discrimination is achieved by learning a multi-class classifier by model distillation and we conduct out-class detection with a binary classifier learned by entropy separation.
We then propose to unify the two subtasks into a self-training framework, regularized by consistency of predictions in local neighborhoods of target samples. Such a learning framework is simple, robust, and optimized easily; it also maintains consistent objective to the theoretical founding of self-training \cite{wei2020theoretical}.
Experiments on several domain adaptation benchmarks demonstrate the superiority of our proposed method over state-of-the-art methods of non-black-box universal domain adaptation. The main contributions of our paper are summarized as follows.
\begin{enumerate}
\item To facilitate practical deployment, we propose a more realistic learning setting of UB$^2$DA, where only the interface of source model is available to the target domain, and where the label-space relations between the two domains are allowed to be different and unknown.

\item To address the challenging UB$^2$DA problem, we unify subtasks of in-class discrimination and out-class detection within a regularized self-training framework, leading to a simple, robust, and easily optimized method. We also illustrate the connection between our method and the existing theoretical result.

\item We show that our simple black-box solution, without access to both source data and source model, significantly outperforms state-of-the-art non-black-box methods in most of the benchmark tasks with the metric of H-score and performs on par with them with metric of averaged class accuracy, justifying its efficacy.  
\end{enumerate}

\section{Related Work}

\subsection{Unsupervised domain adaptation}
Based on different relationships between the label spaces of source and target domains, unsupervised domain adaptation can be divided into two main categories: none-universal and universal domain adaptation. None-universal domain adaptation requires a specific knowledge of relationship of label spaces between source and target domains, including closed set \cite{long2013mmd, tzeng2014mmd, long2015mmd,shen2018wasserstein, chen2019slic-wasserstein,ganin2016dann, tzeng2017adda, saito2018mcd,tang2020-srdc}, partial \cite{cao2018partial-eccv, zhang2018importance,cao2018partial, cao2019partial}, and open set \cite{busto2017openset, saito2018open} domain adaptation. In contrast, universal domain adaptation \cite{you2019universal} assumes that there is no any prior information about the target label space, which thoroughly relaxes the unrealistic assumptions about the label spaces across domains. To solve this problem, You \emph{et al.} \cite{you2019universal} propose the universal adaptation network (UAN) by jointly training an adversarial domain adaptation network and a progressive instance-level weighting scheme, which quantifies the transferability of both source and target samples. Fu \emph{et al.} \cite{fu2020learning} propose calibrated multiple uncertainties (CMU), which is similar to UAN by using instance-level weighting scheme for adversarial alignment, but with a novel weighting solution composed of entropy, consistency, and confidence. Saito \emph{et al.} \cite{saito2020dance} introduce domain adaptative neighborhood clustering (DANCE) via entropy optimization, which shows promising performance on universal domain adaptation. All these seminal works have to require source data for adapting to the target domain, which may violate the privacy concerns in real-world application. Different from these methods, \cite{nath2020UniDA-sourcefree} proposes a novel solution without access to the source data, which is similar to ours, but considers to access to a well-design source model.

Recently, black-box domain adaptation is being concerned due to the widely available off-the-shelf interfaces \cite{nelakurthi2018source} or security considerations \cite{zhang2021unsupervised}. However, these settings are proposed for addressing the closed set problem, which is not realistic since the black-box interface is usually fixed by the provider but our target tasks may be various in different application scenarios. Therefore, it is hard to guarantee that the target categories exactly match that of black-box interfaces. In our learning task, we focus on universal black-box domain adaptation, which is more realitic and much more challenge.

\subsection{Noisy label learning}
Our learning setting is similar to the noisy label learning \cite{angluin1988learning}, which assumes that incorrect labels exist in the training data. There are many ways to address this problem, such as by leveraging the noisy transition matrix \cite{goldberger2016training,patrini2017making,hendrycks2018glc}, by modifying the objective function \cite{azadi2016auxiliary,liu2015classification,wang2017robust,lyu2019curriculum,nguyen2019self}, and by exploiting memorization effects \cite{jiang2018mentornet,han2018co,yu2019does}. However, most of noisy label learning methods cope with specific noise, such as symmetric noise \cite{van2015learning}, asymmetric noise \cite{patrini2017making}, or pure open-set noise \cite{wang2018iterative}. In our universal black-box learning paradigm, we do not make any assumption to the noisy distribution, which means that the true labels of some target noisy data may not be included in the set of known classes of training data. Moreover, our black-box learning setting assumes accessing to the noisy scores of source training classes, which is different from that in noisy label learning as it assumes the noisy labels are one-hot vectors.


\section{Problem Formulation}

Assume a set of unlabeled data $\mathcal{T} = \{\x_i\}_{i=1}^N$ on a target domain $\mathcal{D}_t$ over $\mathcal{X} \times \mathcal{Y}_t$ and a model $f_s: \mathcal{X} \rightarrow \mathbb{R}^{|\mathcal{Y}_s|}$ that was trained on data $\mathcal{S}$ from a source domain $\mathcal{D}_s$ over $\mathcal{X} \times \mathcal{Y}_s$. Both the source data $\mathcal{S}$ and parameters of the model $f_s$ are not available; we also call $f_s$ as the black-box source model. The goal of interest is to learn a target model $h: \mathcal{X} \rightarrow \mathcal{Y}_t$ such that the generalization risk $R(h) = \mathbb{E}_{(X,Y_t)}[l(h(X), Y_t)]$ is minimized, where $l(\cdot,\cdot)$ denotes an appropriate loss function, e.g., 0-1 loss. In this work, we consider the least restrictive setting where $\mathcal{Y}_t$ is unknown and may be different from $\mathcal{Y}_s$, giving rise to the task of universal black-box domain adaptation (UB$^2$DA).

We note that the task of UB$^2$DA can be converted as (1) learning a binary classifier $h_1: \mathcal{X} \rightarrow \{0, 1\}$ to minimize $R_1(h_1) = \mathbb{E}_{(X, Y_t)}[l(h_1(X), I[Y_t \notin \mathcal{Y}_s])]$, where $I[\cdot]$ is an indicator function, and (2) learning a multi-class classifier $h_2: \mathcal{X} \rightarrow \mathcal{Y}_s$ to minimize $R_2(h_2) = \mathbb{E}_{(X, Y_t)}[I[Y_t \in \mathcal{Y}_s]\cdot l(h_2(X), Y_t)]$. As such, learning $h_1$ amounts to detecting those target instances whose labels are not in the label space of $\mathcal{Y}_s$, i.e., \emph{out-class detection}, and learning $h_2$ amounts to discriminating the target instances whose labels are in (possibly a subset of) $\mathcal{Y}_s$, i.e., \emph{in-class discrimination}.

\section{The proposed framework}

In UB$^2$DA, given the black-box source model $f_s$, what we can only have for the target data $\mathcal{T} = \{\x_i\}_{i=1}^N$ are the predictions $\mathcal{F} = \{f_s(\x_i)\}_{i=1}^N$. Due to the inevitable domain gap between $\mathcal{D}_s$ and $\mathcal{D}_t$ (and the possible difference between the label spaces $\mathcal{Y}_t$ and $\mathcal{Y}_s$), each prediction $f_s(\x)$ would be noisy. In spite of being noisy, $\mathcal{F}$ is the only source of information that we can rely on to trigger the learning of $h_1$ and $h_2$. Our overall idea is to decompose the learning objective into those respectively for learning $h_1$ and $h_2$. For the former task, the idea of entropy separation \cite{saito2020dance} has demonstrated its efficacy in the setting of universal domain adaptation \cite{you2019universal}. The later task can be simply solved via model distillation \cite{hinton2015distilling}. For UB$^2$DA studied in the present work, we are motivated by the recent theoretical result \cite{wei2020theoretical}, and propose to unify the aforementioned two learning tasks into a regularized self-training framework. Let $|\mathcal{Y}_s| = K$. We first parameterize $h_1$ and $h_2$ with a function $f: \mathcal{X}\rightarrow \mathbb{R}^{K}$ such that
\begin{eqnarray}\label{Eqnh1h2}
& h_1(\x) = I\left[ H(\sigma(f(\x))) > \log(K)/2 \right]   \\
& h_2(\x) = \argmax_{k}[f(\x)]_k ,
\end{eqnarray}
where $\sigma(\cdot)$ is a softmax function and $H(\cdot)$ is a function to compute entropy. Assuming $f$ has been learned, each target instance $\x$ can be directly classified as
\begin{equation}\label{equ-inference}
h(\x) =
\begin{cases}
h_2(\x) & \text{if} \ h_1(\x)=0, \\
\text{unknown} & \text{if} \ h_1(\x)=1.
\end{cases}
\end{equation}

%

\subsection{A unified self-training}

Since parameters of the source model $f_s$ are not accessible, we initially apply knowledge distillation \cite{hinton2015distilling} to train the network model $f$
\begin{equation}\label{equ-distillation}
\mathcal{L}_{\textrm{\tiny distillation}} = \frac{1}{N}\sum_{i=1}^N CE(\sigma(f_s(\x_i)), \sigma(f(\x_i))),
\end{equation}
where $CE(\cdot,\cdot)$ denotes the cross entropy function. During distillation, the model $f$ would be gradually more discriminative for the target data, which provides a well initialized model for further self-training of the model
\begin{equation}\label{equ-self-train}
\mathcal{L}_{\textrm{\tiny self-training}} = \frac{1}{N}\sum_{i=1}^N g(\x_i)\cdot H(\sigma(f(\x_i))),
\end{equation}
where $g(\cdot)$ is defined as:
\begin{equation}\label{equ-pseudo-label-generator}
g(\x_i) =
\begin{cases}
-1 & \text{if} \ \ H(\sigma(f(\x_i))) > \log(K)/2 + \rho \\
1 & \text{if} \ \ H(\sigma(f(\x_i))) < \log(K)/2 - \rho \\
0 &  \text{otherwise},
\end{cases}
\end{equation}
where $\rho$ is a threshold.

We emphasize that minimizing the loss (\ref{equ-self-train}) has two effects of (1) learning a binary classifier by entropy separation \cite{saito2020dance} and (2) learning a multi-class classifier by entropy minimization. Both of the two effects are equivalent to self-training via pseudo-labeing \cite{Lee2013Pseudolabeling-ssl}. The loss (\ref{equ-self-train}) combing with (\ref{equ-distillation}) is thus a unified self-training objective.


\subsection{Regularization by consistency of predictions in local neighborhoods}
\label{section-self-supervised}

The unified self-training proposed in the preceding section is a general strategy for deep unsupervised learning. However, it is short of considerations common in classical unsupervised or semi-supervised learning, such as minimum coding length \cite{ma2007segmentation,NEURIPS2020_6ad4174e} or density separation \cite{chapelle2005semi}. To improve over self-training, we leverage a simple criterion that promotes the consistency of predictions for target samples in each local neighborhood. Given samples $\{ \bm{x} \}$ in such a neighborhood, we expect the model $f$ to be learned such that elements of $\{ f(\bm{x}) \}$ are consistent. Technically,
we construct the learning model $f$ as a feature extractor $\phi: \mathcal{X}\rightarrow \mathbb{R}^D$ followed by a linear classifier $l:\mathbb{R}^D \rightarrow \mathbb{R}^K$, i.e. $f=l\circ \phi$. We introduce $M$ learnable prototypes $\mathcal{W} = \{\w_1,\w_2,..,\w_M\}$ to be the centers of $M$ clusters of target data features, where $M$ is set to a relatively large number. Then, we search the neighborhoods of each target instance in the feature space via Cosine distance.
Let $p_{im}$ denote the similarity between target feature point $\phi(\x_i)$ and prototype $\w_m$ as
\begin{equation}\label{equ-similarity}
p_{im} = \frac{\text{exp}(d_{im})}{\sum_{m'} \text{exp}(d_{im'})} ,
\end{equation}
with $d_{im}$ the Cosine distance between $\phi(\x_i)$ and prototype $\w_m$, i.e. $d_{im} = \frac{<\phi(\x_i),\w_m>}{\|\phi(\x_i)\|\cdot\|\w_m\|}$. To ensure consistency of neighborhood predictions by $f$ and prevent large clusters from distorting the discriminative feature space, we follow similar strategies of \cite{xie2016unsupervised, ghasedi2017deep} by constructing self-supervised information for each $p_{im}$ as
\begin{equation}\label{equ-auxilary-infomation}
q_{im} = \frac{p_{im}/(\sum_{i'} p_{i'm})^{\frac{1}{2}}}{\sum_{m'} p_{im'}/(\sum_{i'} p_{i'm'})^{\frac{1}{2}}} .
\end{equation}
Then, our regularization of self-supervised style for neighborhood consistency is designed as
\begin{equation}\label{equ-self-supervised-reg}
\mathcal{R}_{\textrm{\tiny regularization}} = \frac{1}{N}\sum_{i=1}^N CE(\q_i, \p_i),
\end{equation}
where $\q_i = [q_{i1}; \dots; q_{iM}]$ and $\p_i = [p_{i1}; \dots, p_{iM}]$. The $M$ prototypes $\mathcal{W} = \{\w_1,\w_2,..,\w_M\}$ are initialized by k-means algorithms to form cluster centers and then updated during the optimization. Once learned, these prototypes are discarded, and only the model $f$ is maintained for inference, as showed in (\ref{equ-inference}).

\subsection{The overall objective}

With both the losses (\ref{equ-distillation}) and (\ref{equ-self-train}) that unify self-training and the regularizer (\ref{equ-self-supervised-reg}), our overall objective of regularized self-training is written as
\begin{align}\label{equ-total-loss}
\mathcal{L} = \alpha\cdot \mathcal{L}_{\textrm{\tiny distillation}} + (1 - \alpha) \cdot \mathcal{L}_{\textrm{\tiny self-training}} + \beta \cdot \mathcal{R}_{\textrm{\tiny regularization}} ,
\end{align}
where $\alpha$ and $\beta$ are penalty parameters.

\section{Theoretical Insight}
Since our technical scheme of regularized self-training framework is motivated by Wei \emph{et al.} \cite{wei2020theoretical}, we want to illustrate the relationship between our technical solution and the theory in this section.

Let $P$ denote a distribution of target examples over input space $\mathcal{X}$. Assume the target data is partitioned into $C$ classes with ground-truth classifier $G^*: \mathcal{X}\rightarrow [C]$. For any $G, G': \mathcal{X}\rightarrow [C]$, define $L_{0-1}(G, G') = \mathbb{E}_{P}[I[G(\x)\neq G'(\x)]]$ to be the disagreement between $G$ and $G'$. Let $Err(G)=L_{0-1}(G, G^*)$ be the expected error of $G$. Denote $G_{pl}$ be the pseudo-labeler and $e_i$ be the fraction of examples in $i$-th class which are mistakenly pseudo-labeled.

Under several conditions, Wei \emph{et al.} \cite{wei2020theoretical} propose the theory that can bound the expected error of a classifier -- trained to fit pseudo labels while regularizing consistency of predictions  in local neighborhoods -- by a term that is smaller than the error of pseudo-labeler. This theory can be explained as follows.
\begin{theorem}\label{theorem}
Suppose $P$ satisfies expansion property (Assumption 4.1 in \cite{wei2020theoretical}). Then for any minimizer $\hat{G}$ of
\begin{align}\label{equ-theory-2}
\min_G \mathcal{L}(G) = \max\{2R_\mathcal{B}(G) + L_{0-1}(G, G_{pl}) - Err(G_{pl}), \notag \\
4R_\mathcal{B}(G) + 3L_{0-1}(G, G_{pl}) - (3-\frac{4}{c-1})Err(G_{pl}) \notag\},
\end{align}
we have
\begin{equation}\label{equ-theory-1}
Err(\hat{G}) \leq \frac{4}{c-1} Err(G_{pl}) + 4R_{\mathcal{B}}(G^*) \notag
\end{equation}
where
\begin{align}
R_{\mathcal{B}}(G) = \mathbb{E}_{P}[I[\exists \x'\in\mathcal{B}(\x) \ \text{such that} \ G(\x')\neq G(\x)]] \notag
\end{align}
Here, $\mathcal{B}(\x)$ can be considered as the neighborhood around $\x$, $R_{\mathcal{B}}(G)$ represents the opposite degree of neighborhood consistency by $G$, and $c>\frac{1}{\max_i{e_i}}$ in Theorem \ref{theorem} is a value that represents expansion degree of population $P$.
\end{theorem}

Assume $R_\mathcal{B}(G^*)$ is very small and approach to zero, the above theorem shows that under the expansion assumption, if we have a pseudo-laber with $\max_i e_i < 0.2$, then $c>5$ and we can get $\hat{G}$ by minimizing $\mathcal{L}(G)$ such that $Err(\hat{G})< Err(G_{pl})$.

In our method, we can consider that there are two self-training tasks for minimizing $R_1(h_1)$ and $R_2(h_2)$, corresponding to minimizing $Err(G)$ with $C=2$ and $C=K$ respectively. Minimizing $\mathcal{L}_{\textrm{\tiny self-training}}$ loss in (\ref{equ-self-train}) is equivalent to apply pseudo-labeling to fit the pseudo-labels, which corresponds to minimizing the term $L_{0-1}(G,G_{pl})$ in Theorem \ref{theorem}. Our distillation loss $\mathcal{L}_{\textrm{\tiny distillation}}$ and the designed strategy of (\ref{equ-pseudo-label-generator}) are to ensure that the pseudo-labels are mostly correct such that $\max_i e_i$ is small enough, and the designed regularization of $\mathcal{R}_{\textrm{\tiny regularization}}$ is to match the goal of neighborhood consistency of predictions by both $h_1$ and $h_2$, which demonstrates the similar effect of minimizing $R_\mathcal{B}(G)$ in Theorem \ref{theorem} . Therefore, minimizing $\mathcal{L}$ in (\ref{equ-total-loss}) is approximately equivalent to minimize $\mathcal{L}(G)$ in Theorem \ref{theorem} for the both two tasks. Assume the conditions in Theorem \ref{theorem} hold and the pseudo-labeler initialized from the source black-box model $f_s$ by distillation satisfies $\max_i e_i < 0.2$ for both the two classification tasks, then we can get decreasing errors of both $R_1(h_1)$ and $R_2(h_2)$ during the iterative self-training process.

\section{Experiment}
In this section, we evaluate our method on three universal domain adaptation benchmarks under the UB$^2$DA setting.

\begin{table*}[t]
\begin{center}
\begin{small}
\begin{tabular}{clccccccr}
\toprule
\multicolumn{9}{c}{H-score (\%)} \\
\midrule
Setting & Method           &    A2W    &    D2W    &    W2D    &    A2D    &    D2A    &    W2A    &    Avg.    \\
\midrule
\multirow{3}{*}{Non-Black-Box} & UAN \cite{you2019universal}             &    58.6   &    70.6   &    71.4   &    59.7   &    60.1   &    60.3   &    63.5    \\

& CMU \cite{fu2020learning}              &    67.3   &    79.3   &    80.4   &    68.1   &    71.4   &    72.2   &    73.1    \\

& DANCE \cite{saito2020dance}           &    67.4   &    89.9   &    \textbf{90.7}   &    70.8   &    79.1   &    71.9   &    78.3    \\

\midrule
\multirow{2}{*}{Black-Box} & SO++             &    50.3   &    74.9   &    61.5   &    48.1   &    77.0   &    70.8   &    63.8    \\
 & Ours    &    \textbf{78.2}   &    \textbf{92.6}   &    87.9   &    \textbf{80.9}   &    \textbf{92.6}   &    \textbf{89.4}   &    \textbf{86.8}    \\
\midrule
\multicolumn{9}{c}{AA (\%)} \\
\midrule
%
%
%
%
%
%

\multirow{4}{*}{Non-Black-Box} & UAN \cite{you2019universal}              &    86.6   &    94.8   &    98.0   &    86.5   &    85.5   &    85.1   &    89.2    \\

& USFDA \cite{nath2020UniDA-sourcefree}           &    85.6   &    95.2   &    97.8   &    88.5   &    87.5   &    86.6   &    90.2    \\

& CMU \cite{fu2020learning}             &    86.9   &    95.7   &    98.0   &    89.1   &    88.4   &    88.6   &    91.1    \\

& DANCE \cite{saito2020dance}            &    \textbf{92.8}   &    97.8   &    \textbf{97.7}   &    \textbf{91.6}   &   \textbf{92.2}   &   \textbf{91.4}   &    \textbf{93.9}    \\
\midrule
\multirow{2}{*}{Black-Box} & SO++             &    75.0   &    94.1   &    94.6   &    83.0   &    81.2   &    83.6   &    85.3    \\
& Ours    &    83.0   &    \textbf{98.1}   &    97.8   &    88.7   &    91.4   &    91.0   &    91.7    \\
\bottomrule
\end{tabular}
\end{small}
\end{center}
\caption{Classification results of tasks on \textbf{Office31} dataset}
\label{table-Office31-OPDA-HScore}
\end{table*}

%
%
%

\begin{table*}[t]
\begin{center}
\begin{small}
\begin{tabular}{clccccccccccccr}
\toprule
\multicolumn{15}{c}{H-score (\%)} \\
\midrule
Setting & Method           &  A2C  &  A2P  &  A2R  &  C2A  &  C2P  &  C2R  &  P2A  &  P2C  &  P2R  &  R2A  &  R2C  &  R2P  &  Avg. \\
\midrule
\multirow{3}{*}{Non-Black-Box} & UAN \cite{you2019universal}              &  51.6 &  51.7 &  54.3 &  61.7 &  67.6 &  61.9 &  50.4 &  47.6 &  61.5 &  62.9 &  52.6 &  65.2 &  56.6 \\

& CMU \cite{fu2020learning}              &  56.0 &  56.9 &  59.2 &  67.0 &  64.3 &  67.8 &  54.7 &  51.1 &  66.4 &  68.2 &  57.9 &  69.7 &  61.6 \\

& DANCE \cite{saito2020dance}            &  35.9 &  29.3 &  35.2 &  42.6 &  18.0 &  29.3 &  50.2 &  45.4 &  41.1 &  19.8 &  38.8 &  52.6 &  36.5 \\

\midrule
\multirow{2}{*}{Black-Box} & SO++             &  49.8 &  50.1 &  53.2 &  62.4 &  50.3 &  57.4 &  61.1 &  49.5 &  56.1 &  56.9 &  53.0 &  53.7 &  54.5 \\
& Ours    &  \textbf{60.9} &  \textbf{69.6} &  \textbf{76.3} &  \textbf{74.4} &  \textbf{69.2} &  \textbf{76.5} &  \textbf{74.5} &  \textbf{60.3} &  \textbf{76.2} &  \textbf{74.1} &  \textbf{62.0} &  \textbf{71.1} &  \textbf{70.4}\\
\midrule
\multicolumn{15}{c}{AA (\%)} \\
\midrule
%
%

\multirow{3}{*}{Non-Black-Box} & UAN \cite{you2019universal}             &  63.0 &  82.8 &  87.9 &  76.9 &  78.7 &  85.4 &  78.2 &  58.6 &  86.8 &  83.4 &  63.2 &  79.4 &  77.0 \\

& USFDA \cite{nath2020UniDA-sourcefree}           &  63.4 &  83.3 &  89.4 &  71.0 &  72.3 &  86.1 &  78.5 &  60.2 &  87.4 &  81.6 &  63.2 &  88.2 &  77.0 \\

& DANCE \cite{saito2020dance}           &  \textbf{64.1} &  84.3 &  91.2 &  \textbf{84.3} &  \textbf{78.3} &  89.4 &  \textbf{83.4} &  \textbf{63.6} &  91.4 &  \textbf{83.3} &  \textbf{63.9} &  86.9 &  \textbf{80.4} \\
\midrule
\multirow{2}{*}{Black-Box} & SO++             &  55.4 &  81.0 &  91.0 &  71.2 &  71.0 &  83.6 &  70.3 &  50.9 &  89.5 &  77.6 &  56.6 &  84.7 &  63.8 \\
& Ours    &  56.8 &  \textbf{86.9} &  \textbf{94.4} &  70.8 &  76.4 &  \textbf{91.2} &  77.0 &  56.6 &  \textbf{93.0} &  82.0 &  58.5 &  \textbf{89.2} &  77.7\\
\bottomrule
\end{tabular}
\end{small}
\end{center}
\caption{Classification results of tasks on \textbf{Office-Home} dataset}
\label{table-Officehome-OPDA-HScore}
\end{table*}

\begin{table*}[t]
\begin{center}
\begin{small}
\begin{tabular}{clccccccc}
\toprule
Setting & Method   &   P2R    &    R2P    &    P2S    &    S2P    &    R2S    &    S2R    &    Avg.  \\
\midrule
\multirow{3}{*}{Non-Black-Box} & UAN \cite{you2019universal}            &  41.85 &  43.59 &  39.06 &  38.95 &  38.73 &  43.69 &  40.98  \\ 
 & CMU \cite{fu2020learning}            &  50.78 &  52.16 &  45.12 &  \textbf{44.82} &  45.64 &  50.97 &  48.25  \\ 
 & DANCE \cite{saito2020dance}  &    35.15    &    49.24    &    43.32    &    40.18    &    \textbf{46.24}    &     36.57   &    41.78  \\ 
 \hline
\multirow{2}{*}{Black-Box}  & SO++             &    47.30    &    48.19    &    43.89    &    35.93    &    41.99    &    43.71    &    43.50   \\
 & Ours    &    \textbf{57.10}    &    \textbf{54.76}    &    \textbf{47.23}    &    41.39    &    44.03    &    \textbf{51.53}    &    \textbf{49.31}  \\
\bottomrule
\end{tabular}
\end{small}
\end{center}
\caption{Classification results (\textbf{H-score (\%)}) of universal domain adaptation on \textbf{DomainNet} dataset. Note that all comparison methods did not report the results of AA for this dataset in their papers.}
\label{table-domainNet-OPDA-HScore}
\end{table*}

\subsection{Setup}
\noindent \textbf{Datasets.} As the most previous used datasets, \textbf{Office31} \cite{office31}, \textbf{Office-Home} \cite{officehome}, and a large scale one \textbf{DomainNet} \cite{peng2019moment} are used in our experiments.
For all datasets, we follow the same setup as \cite{fu2020learning}.
Specifically, Office31 has 31 classes shared by three domains of Amazon (A), Dslr (D), and Webcam (W). The 10 classes shared by Office31 and Caltech-256 are used as the common label set between source and target domains and then in alphabetical order, the next 10 classes and the remaining 11 classes are used as the source private classes and the target private classes respectively. Office-Home is a larger dataset containing 65 classes and four domains of Artistic (A), Clip-Art (C), Product (P), and Real-World (R). In alphabetical order, the first 10 classes are selected as the common classes, the next 5 classes as the source private classes, and the rest as target private classes.
DomainNet \cite{peng2019moment} is the largest domain adaptation dataset by far, which contains six domains: Clipart(C), Infograph(I), Painting(P), Quickdraw(
Q), Real(R) and Sketch(S) across 345 classes. Following \cite{fu2020learning}, three domains of P, R, and S are selected in the experiments. In alphabetical order, we use the first 150, next 50, and remaining classes as the common classes, source and target private ones, respectively, as the same setup as \cite{fu2020learning}.

\noindent \textbf{Evaluation protocols.} As pointed out in \cite{fu2020learning}, previous works \cite{you2019universal, saito2020dance} treated all out-class classes as an extra class and used \textbf{Average Class Accuracy (AA)} for evaluation can not truly reflect the ability of the algorithm for out-class detection as it is badly biased to accuracy of in-class classes. In our experiments, we follow the same evaluation protocol of \cite{fu2020learning} by using \textbf{H-score} as the main evaluation metric. The H-score is defined as:
\begin{equation}
h = 2 \cdot \frac{Acc_{in} \cdot Acc_{out}}{Acc_{in} + Acc_{out}},
\end{equation}
where $Acc_{in}$ and $Acc_{out}$ represent the instance accuracy on in-class and out-class categories respectively. This score is high only when both tasks of in-class discrimination and out-class detection are well performed. Besides, we also present AA for more comprehensive comparisons to previous methods.

\noindent \textbf{Compared methods.} \textbf{Source-only (SO)} is the model trained only on the source data without using target data. However, SO can not apply for out-class detection. We modify it to \textbf{SO++} by using the same inference as ours as (\ref{equ-inference}), which builds a baseline for universal domain adaptation. Besides SO++, we compare our method mainly to the previous universal domain adaptation approaches, including \textbf{UAN} \cite{you2019universal}, \textbf{DANCE} \cite{saito2020dance}, \textbf{CMU} \cite{fu2020learning}, and \textbf{USFDA} \cite{nath2020UniDA-sourcefree}. Note that these methods all need to access to source data or source model for adapting and without work in our black-box setting.

\noindent \textbf{Implementation.} All experiments are implemented in Pytorch \cite{paszke2017automatic} with Stochastic Gradient Descent (SGD) optimizer. For fair comparisons to previous methods, we use the same backbone of ResNet50 \cite{he2016deep} pre-trained on ImageNet to obtain the source black-box model, which was fine-tuned on source examples optimizing with cross-entropy loss function and then treat it like a black-box by only requiring the input-output interfaces of this model in our experiments. For the target model, we use the same backbone as the comparing methods of pre-trained ResNet50 but we also analyse the performance of our method when using different backbones. We fix the hyperparameter of $\rho$ to 0.5 in all our experiments. For the number of prototypes $M$ and weight parameter $\beta$, we set them to 100 and 1.0 for small scale datasets: office31 and officehome, and to 1000 and 0.05 for the large scale dataset: DomainNet. The analysis of sensitivity of these hyperparameters are discussed. The initial value of $\alpha$ is set as 1, which is decayed with the factor of $1-\frac{t}{100}$, where $t$ denotes the number of epoches. We run each experiment for 100 epochs and report the average result over three random runs. The code and the learning settings of our experiments are available in supplementary.

\begin{figure}[t]
\begin{center}
   \includegraphics[width=0.8\linewidth]{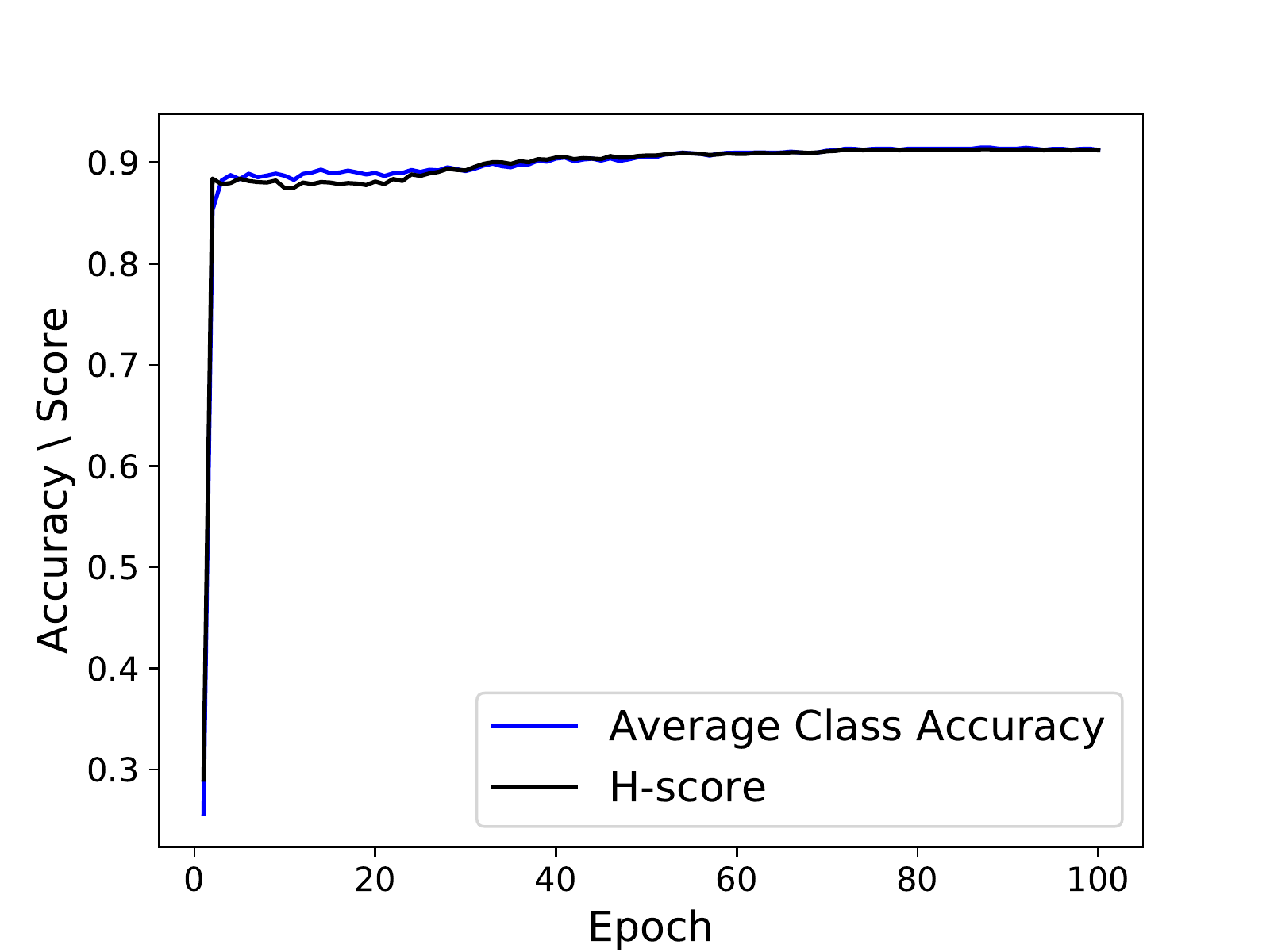}
\end{center}
   \caption{Convergence analysis in D2A task. (Best viewed in color)}
\label{fig-convergence}
\end{figure}

\begin{figure}[t]
  \centering
  \subfigure[Source Model]{
     \label{fig-tsne-so}
     \includegraphics[width=0.4\linewidth]{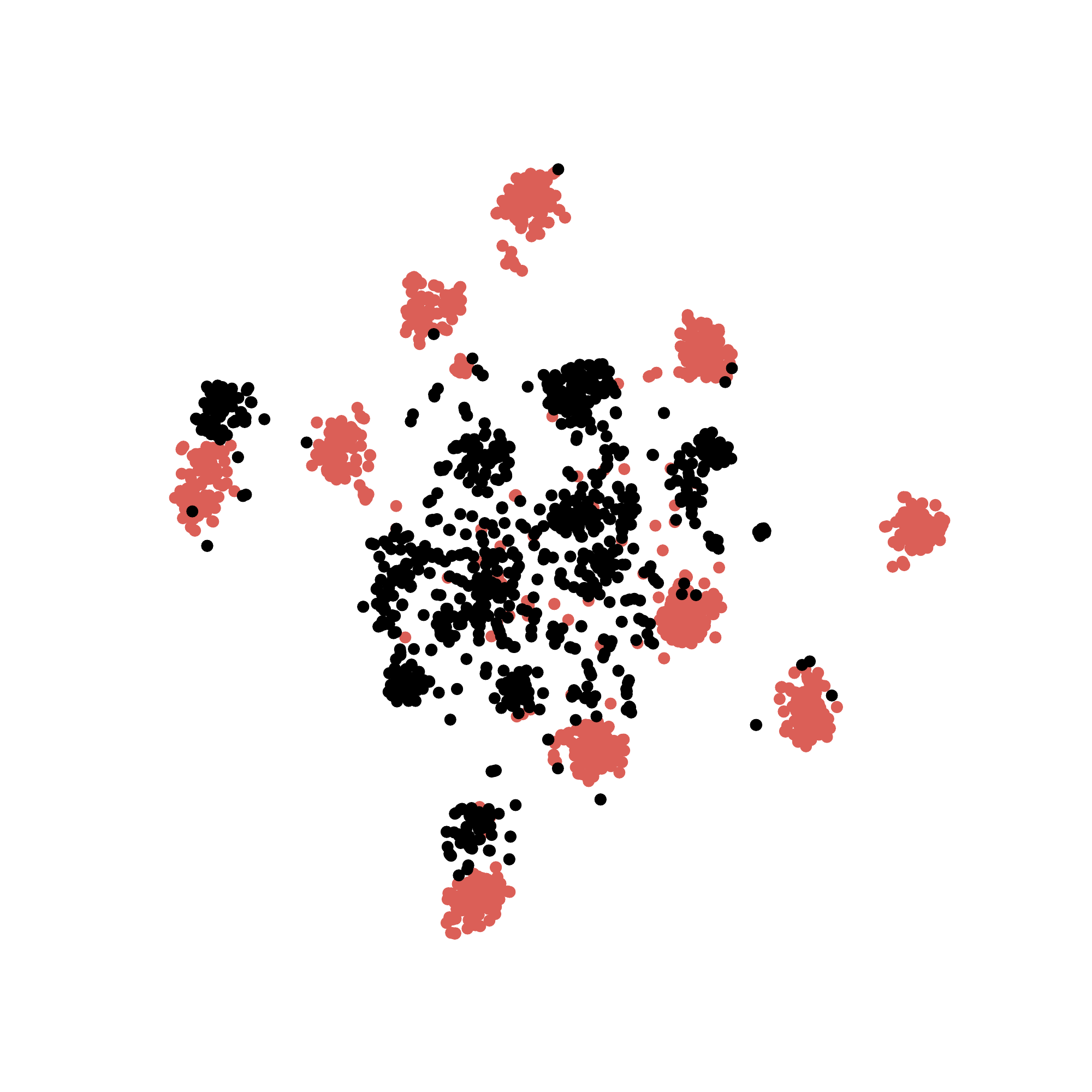}}
  \subfigure[Ours]{
     \label{fig-tsne-ours}
     \includegraphics[width=0.4\linewidth]{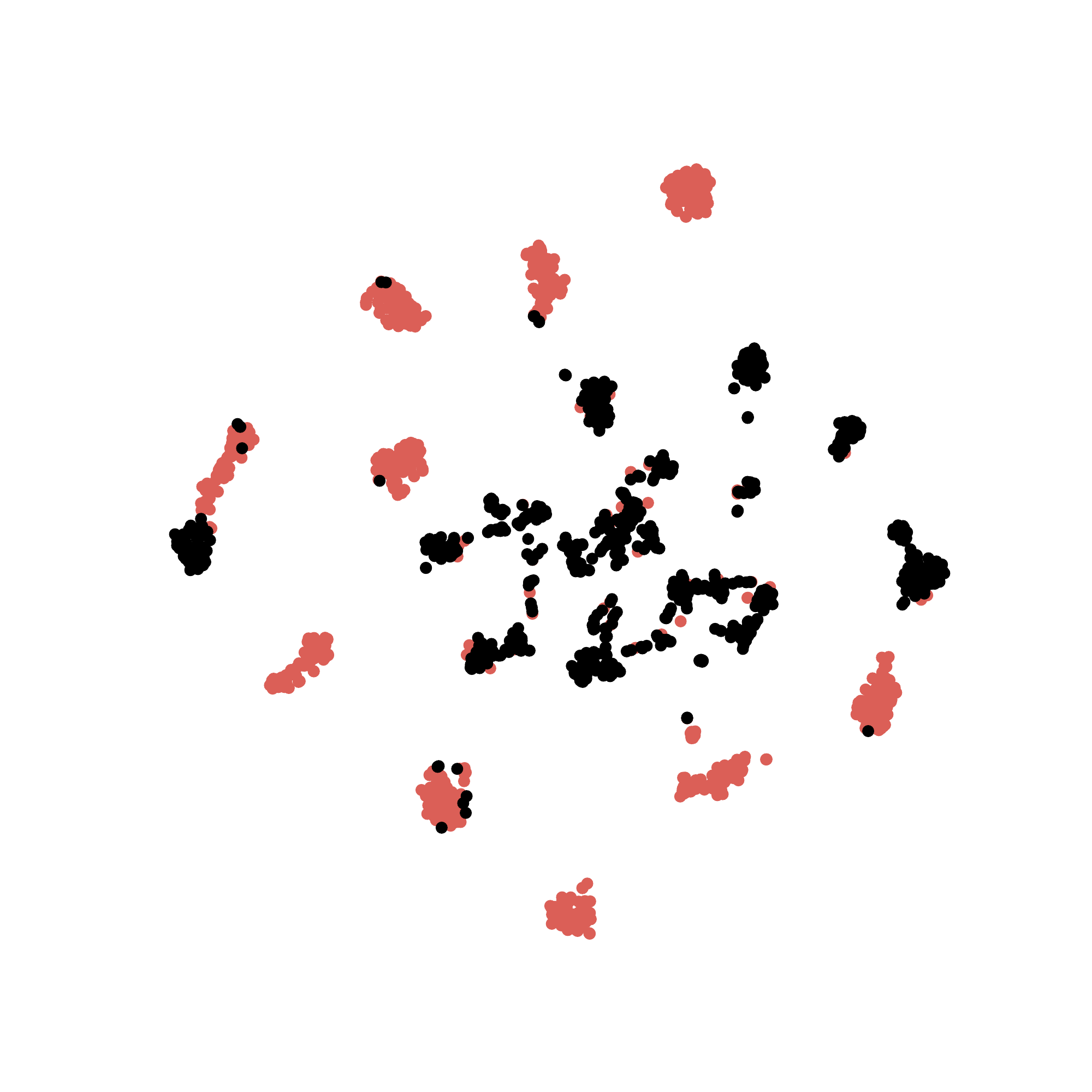}}
  \caption{t-SNE feature visualization of target representations in D2A task. Red dots represent in-class target examples while black dots represent out-class examples. (Best viewed in color)}\label{fig-tsne}
\end{figure}

\begin{table*}[t]
\begin{center}
\begin{small}
\begin{tabular}{ccc|cccc}
\toprule
\multirow{2}{*}{$\mathcal{L}_{\textrm{\tiny distillation}}$} & \multirow{2}{*}{$\mathcal{L}_{\textrm{\tiny self-training}}$} & \multirow{2}{*}{$\mathcal{R}_{\textrm{\tiny regularization}}$} & \multicolumn{2}{c}{A2W} & \multicolumn{2}{c}{W2A} \\
\cmidrule(r){4-5} \cmidrule(r){6-7}  & & & AA & H-score & AA & H-score \\ \hline
$\checkmark$ & - & - & 76.9 & 67.5 & 83.3 & 76.6 \\
$\checkmark$ & \checkmark & - & 80.6 & 77.6 & 88.6 & 88.0 \\
$\checkmark$ & \checkmark & \checkmark & 83.0 & 78.2 & 91.0 & 89.4 \\ \hline
\multicolumn{2}{c}{SO++} &  & 75.0 & 50.3 & 83.6 & 70.8 \\
\bottomrule
\end{tabular}
\end{small}
\end{center}
\caption{Ablation Study. $\mathcal{L}_{\textrm{\tiny distillation}}$, $\mathcal{L}_{\textrm{\tiny self-training}}$, and $\mathcal{R}_{\textrm{\tiny regularization}}$ represent the loss functions defined in equations (\ref{equ-distillation}), (\ref{equ-self-train}), and (\ref{equ-self-supervised-reg}) respectively. These three functions constitute our final optimization objective (\ref{equ-total-loss}).}
\label{table-ablation-study}
\end{table*}

\subsection{Results}
Tables \ref{table-Office31-OPDA-HScore}, \ref{table-Officehome-OPDA-HScore}, and \ref{table-domainNet-OPDA-HScore} show the classification results on Office31, Office-Home, and DomainNet datasets.
We can see from the figure that the H-score of UAN and DANCE methods even worse that the baseline of SO++ in some of tasks even though their results of AA are good. The reason behind this is because these two methods perform badly on out-class detection due to the overemphasis on alignment between source and target domains. However, our method, learning on a black-box setting, significantly outperforms previous non-black-box universal domain adaptation methods with respect to the H-score and show a large margin improvement to the baseline of SO++ for all tasks. Interestingly, in terms of the AA, although our method is inferior to DANCE, it still outperforms other non-black-box methods such as UAN \cite{you2019universal}, USFDA \cite{nath2020UniDA-sourcefree}, and CMU \cite{fu2020learning} on Office31 and Office-Home datasets, which demonstrates clear advantages of our black-box method.

\subsection{Analysis}
\noindent \textbf{Ablation study.} Here, we conduct ablation studies to demonstrate the effectiveness of each loss component in (\ref{equ-total-loss}). The results are illustrated in Table \ref{table-ablation-study}. Firstly, comparing to the baseline of SO++, where the results are directly taken from the output of the black-box source model, target model by using only distillation for training shows large improvement on H-score while almost remaining the same on AA. This demonstrates that distillation from source to target could prevent over-confident predictions, leading to better out-class detection. After gradually adding the self-training loss $\mathcal{L}_{\textrm{\tiny self-training}}$ and neighborhood consistency regularization $\mathcal{R}_{\textrm{\tiny regularization}}$ to the optimization, we can see further improvement on both AA and H-score, verifying the effectiveness of the self-training module for addressing our learning task and the usefulness of neighborhood consistency constraint for improving the self-training, which is consistent with our theoretical analysis.

\noindent \textbf{Convergence analysis.} Since our algorithm is trained in an end-to-end manner with SGD optimization, we show its convergence performance with respect to the classification results of H-score and AA on the task of D2A, as illustrated in Figure \ref{fig-convergence}. It can be seen that our algorithm converge very fast and show almost gradual upward trend on both H-score and AA, showing the stability of our proposed method.

\noindent \textbf{Feature visualization.} We visualize in Figures \ref{fig-tsne-so} and \ref{fig-tsne-ours} the output representations extracted by source model as well as ours on the D2A task by t-SNE \cite{maaten2008visualizing}. Compared to source model, our method learns more discriminative representations for the target in-class examples and show clear separation between the in-class and out-class examples.

\begin{figure*}[t]
\begin{center}
   \subfigure[AA w.r.t $|\mathcal{Y}_s-\mathcal{Y}|$]{
     \label{fig-openness-acc}
     \includegraphics[width=0.24\textwidth]{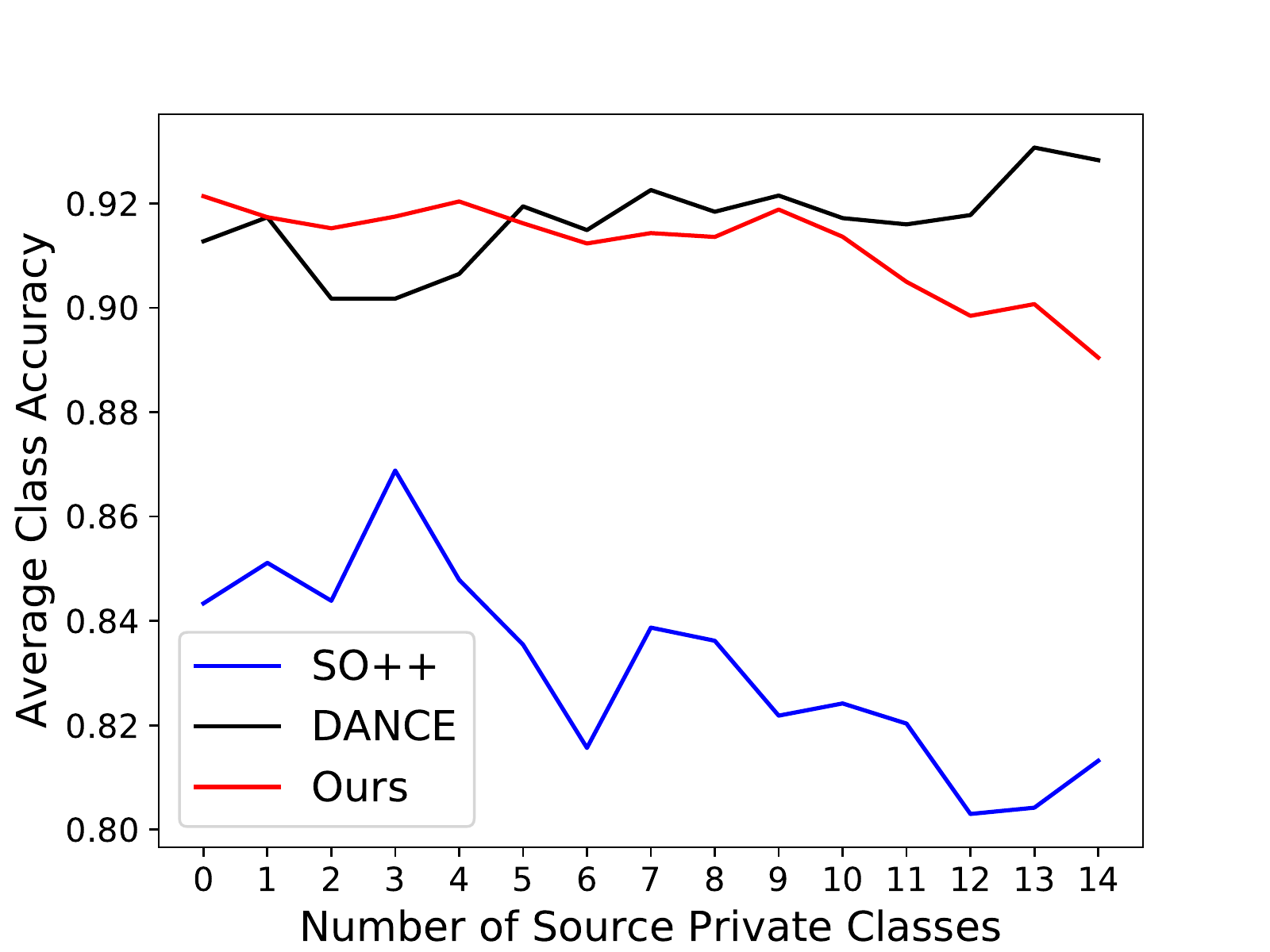}}
  \subfigure[H-score w.r.t $|\mathcal{Y}_s-\mathcal{Y}|$]{
     \label{fig-openness-hscore}
     \includegraphics[width=0.24\textwidth]{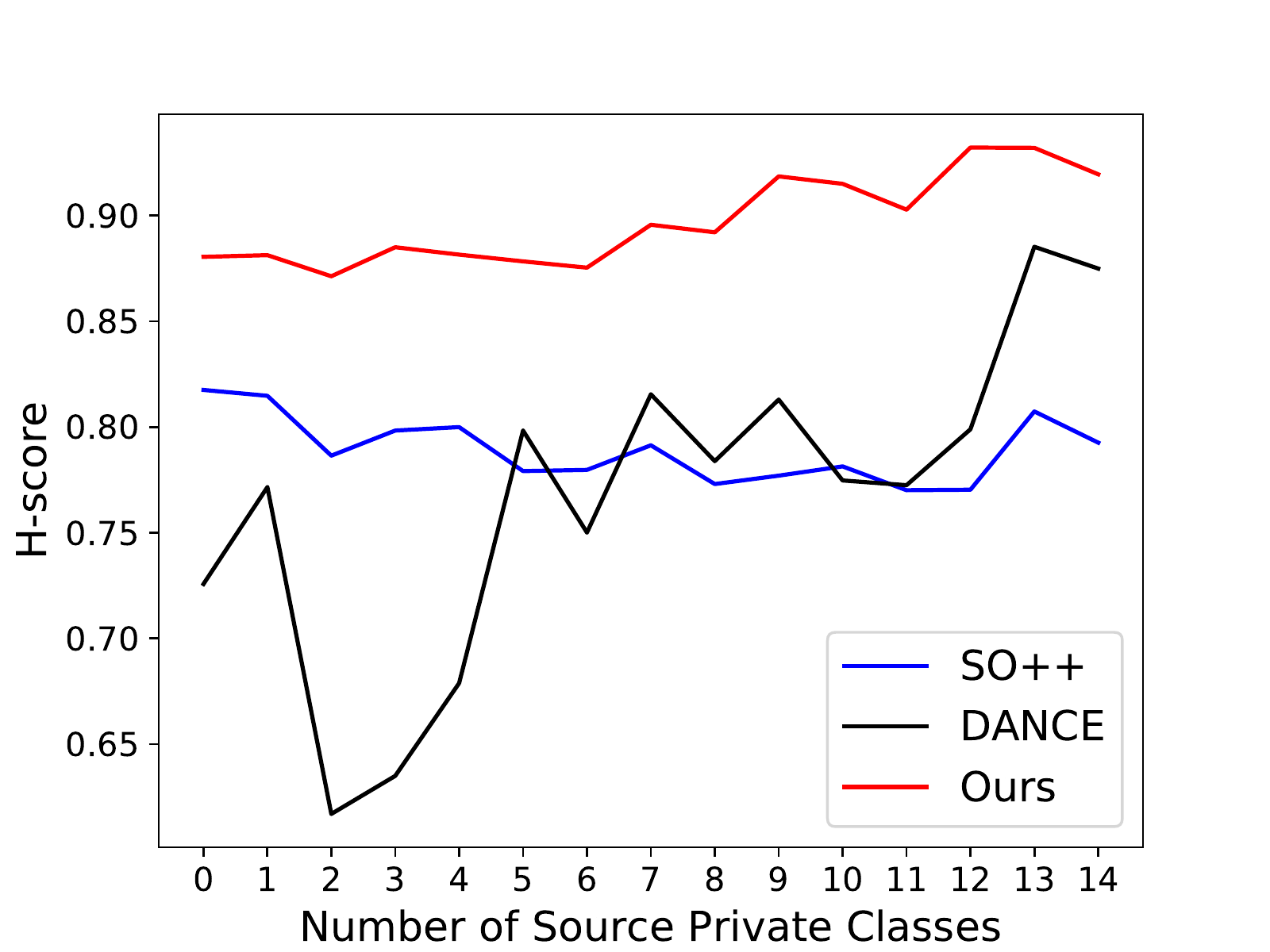}}
  \subfigure[AA w.r.t $|\mathcal{Y}|$]{
     \label{fig-openness-acc-v2}
     \includegraphics[width=0.24\textwidth]{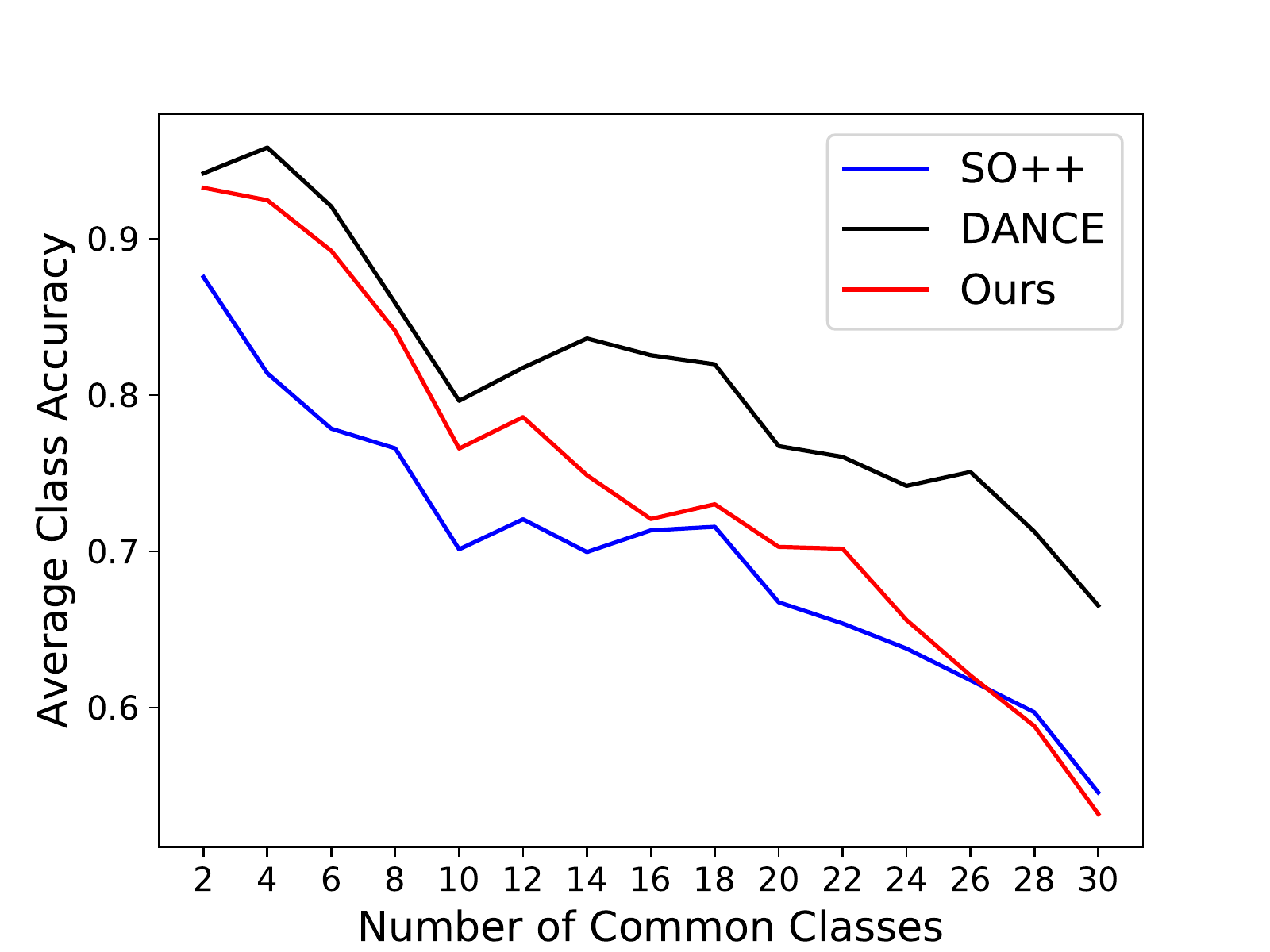}}
  \subfigure[H-score w.r.t $|\mathcal{Y}|$]{
     \label{fig-openness-hscore-v2}
     \includegraphics[width=0.24\textwidth]{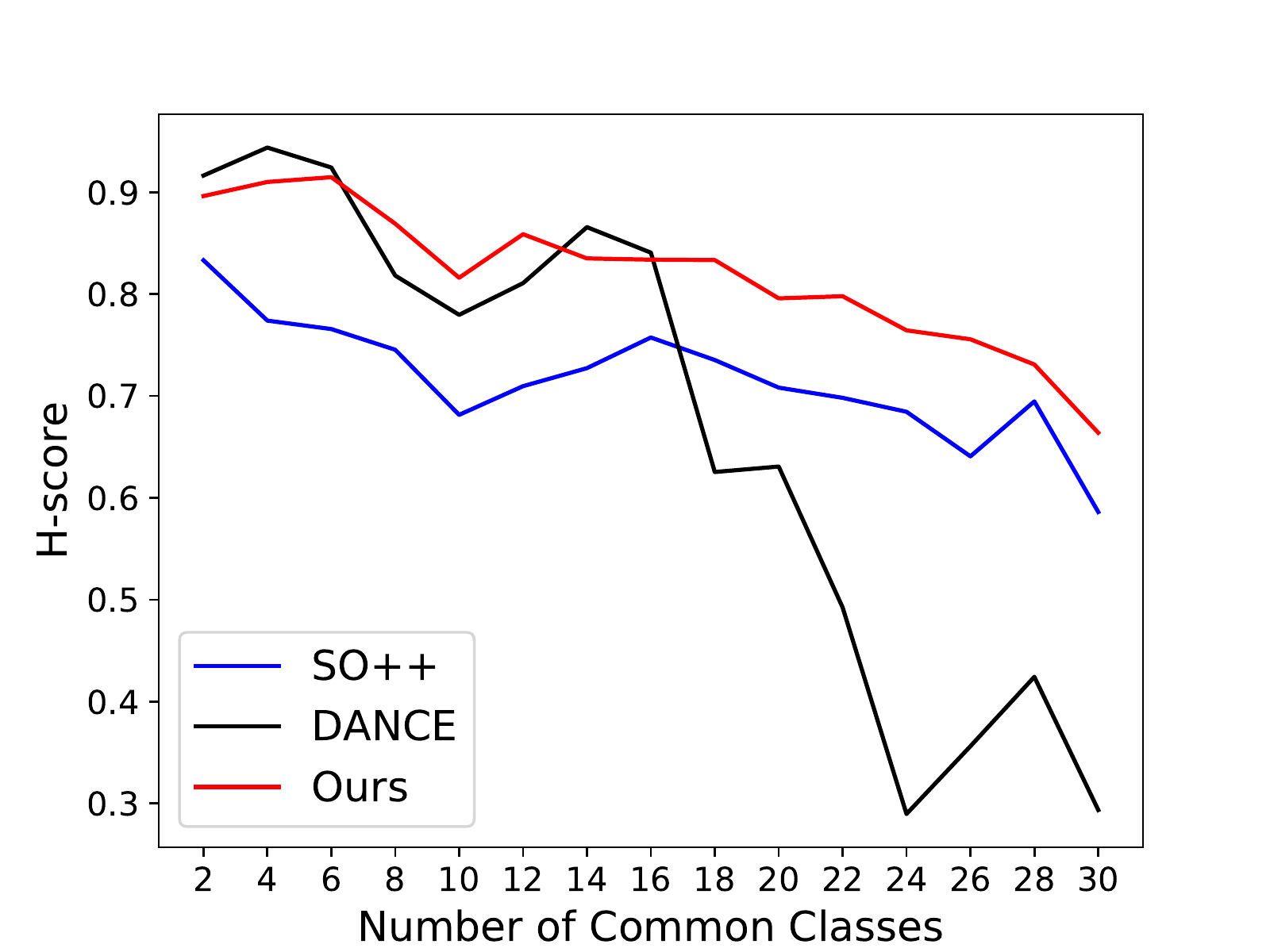}}
\end{center}
   \caption{Average Class Accuracy and H-score with respect to the number of source private classes of $|\mathcal{Y}_s-\mathcal{Y}|$ and the number of common classes of $|\mathcal{Y}|$, where $\mathcal{Y} = \mathcal{Y}_s\cap\mathcal{Y}_t$. In (a) and (b), we fix $|\mathcal{Y}|=10$ and $|\mathcal{Y}_s\cup\mathcal{Y}_t|=31$; In (c) and (d), we fix $|\mathcal{Y}_s| + 1 = |\mathcal{Y}_t|$ and $|\mathcal{Y}_s\cup\mathcal{Y}_t|=31$. Note that the strategy of selecting sets $\mathcal{Y}$, $\mathcal{Y}_s-\mathcal{Y}$, and $\mathcal{Y}_t-\mathcal{Y}$ is different among these two settings. In the first setting, the fixed $\mathcal{Y}$ is the class set shared between Caltech256 and Office31 and the remaining two sets are selected in alphabetical order, while in the second setting, we select them correspondingly in alphabetical order. (Best viewed in color)}
\label{fig-openness}
\end{figure*}

\noindent \textbf{Analysis on different categories relationship between source and target domains.} Figure \ref{fig-openness} shows the classification results on D2A task under different degrees of overlap categories between source and target domains. Here, the currently state-of-the-art method of DANCE \cite{saito2020dance} and the baseline of SO++ are compared with ours. Denote $\mathcal{Y}=\mathcal{Y}_s \cap \mathcal{Y}_t$ be the common label set between source and target domains and $|\mathcal{Y}|$ be the number of classes in $\mathcal{Y}$. Firstly, we set $|\mathcal{Y}|=10$ with $\mathcal{Y}$ be the classes shared between Caltech256 and Office31 datasets and vary $|\mathcal{Y}_s-\mathcal{Y}|$ to construct different source and target domains ($|\mathcal{Y}_t|$ therefore changes correspondingly since $|\mathcal{Y}_s\cup\mathcal{Y}_t| = 31$). In this case, we range the remaining 21 classes to $\mathcal{Y}_s - \mathcal{Y}$ and $\mathcal{Y}_t-\mathcal{Y}$ in alphabetical order respectively. The running results of AA and H-score under this setting are illustrated in Figures \ref{fig-openness-acc} and \ref{fig-openness-hscore} respectively. Second, we vary the number of $|\mathcal{Y}|$ and let $|\mathcal{Y}_s-\mathcal{Y}| + 1 = |\mathcal{Y}_t-\mathcal{Y}|$ to construct different source and target domains, correspondingly showed in Figures \ref{fig-openness-acc-v2} and \ref{fig-openness-hscore-v2}. For that, we select the class set $\mathcal{Y}$, $\mathcal{Y}_s-\mathcal{Y}$, and $\mathcal{Y}_t-\mathcal{Y}$ in alphabetical order correspondingly. From these figures, we can see that our method consistently outperforms the baseline of SO++ across all scenarios, and improves the state-of-the-art of DANCE in most of tasks with a large margin with respect to the H-score. Although our method do not show much advantage related to the AA while comparing to the DANCE, it still significantly better than the baseline of SO++ overall.

\begin{figure*}[ht]
  \centering
  \subfigure[]{
     \label{fig-margin}
     \includegraphics[width=0.25\textwidth]{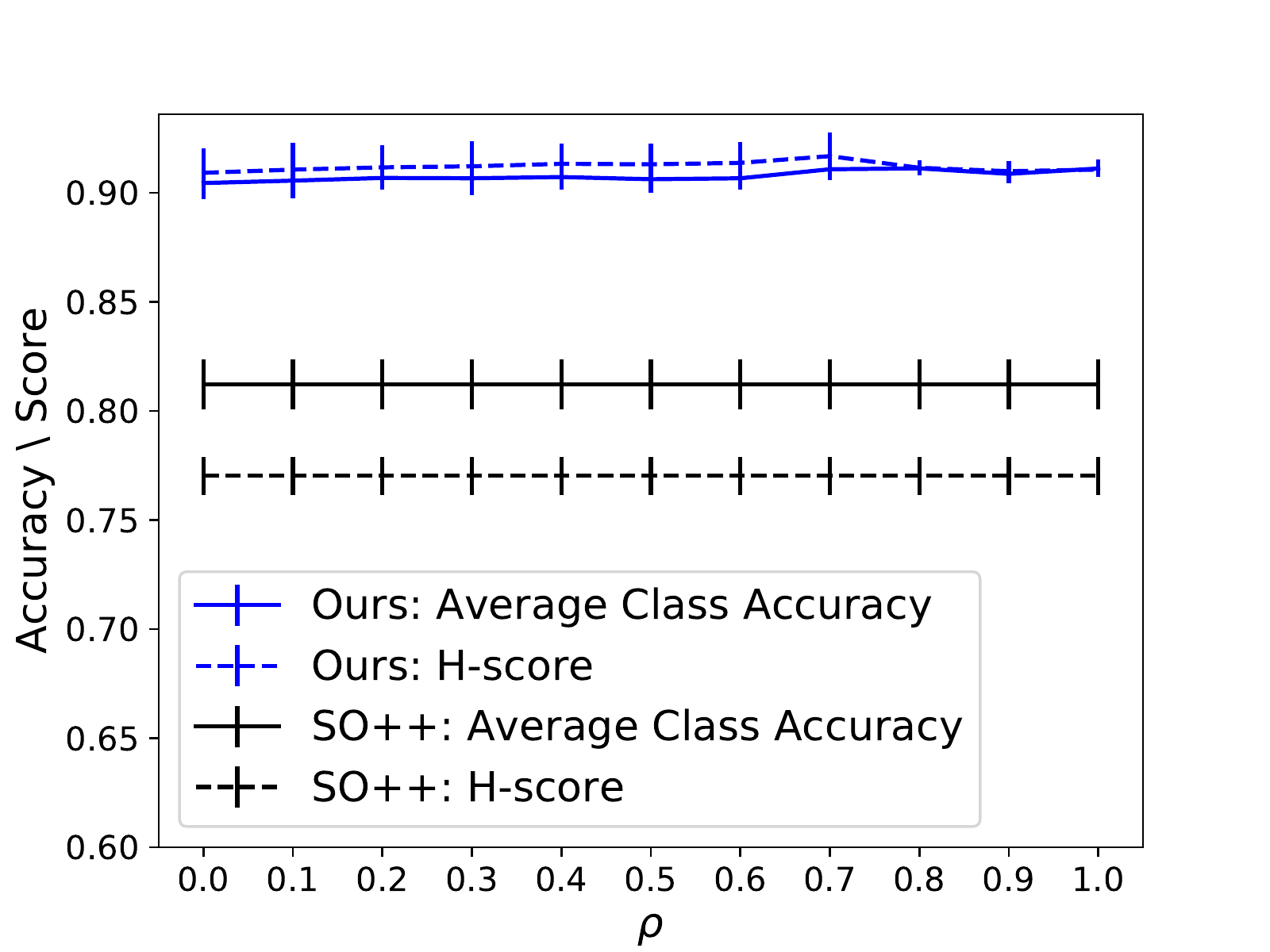}}
  \subfigure[]{
     \label{fig-beta}
     \includegraphics[width=0.25\textwidth]{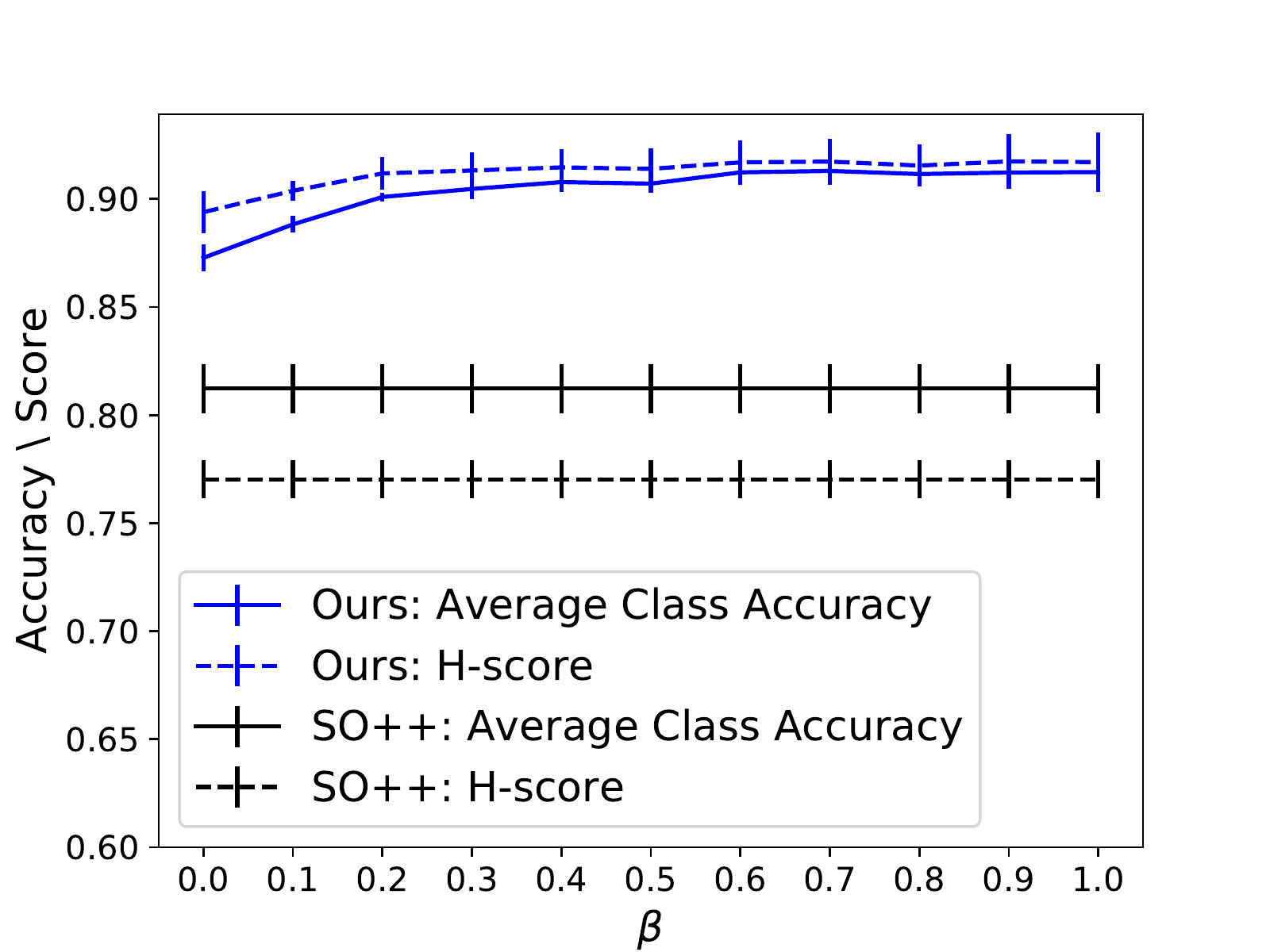}}
  \subfigure[]{
     \label{fig-clusterNum}
     \includegraphics[width=0.25\textwidth]{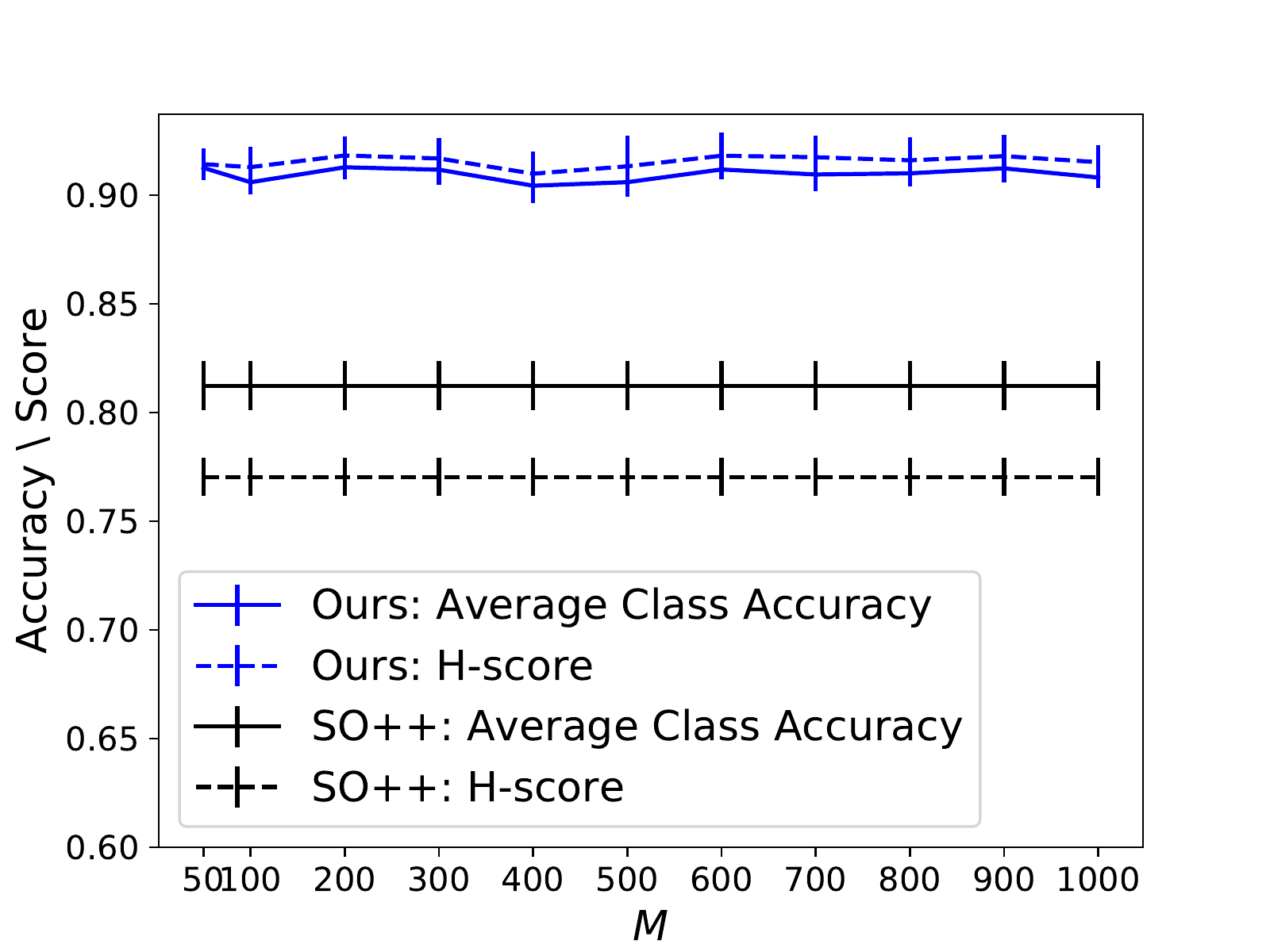}}
  \caption{(a)-(c): Parameters sensitive analysis in D2A task, where $\rho$ is the threshold in (\ref{equ-pseudo-label-generator}), $\beta$ is the weight parameter in (\ref{equ-total-loss}), and $M$ is the number of prototypes defined in Section \ref{section-self-supervised}. The default values of these hyperparameters are set to $\rho=0.5$, $\beta=1.0$, and $M=100$ in this task. (a), (b), and (c) respectively show the effect of changing one of the parameters on the result. Note that we run each experiment three random times and report the average results and standard deviations in the figure, including that in SO++. (Best viewed in color)}\label{fig-parameter-analysis}
\end{figure*}

\noindent \textbf{Parameters sensitivity analysis.} To analyse how hyperparameters of $\rho$, $\beta$, and $M$ affect our experimental results, we run each experiment three times on D2A task across a wide range of different values of each hyperparameter while fixing others as default and report their average results and standard deviations in Figure \ref{fig-parameter-analysis}. From the presented results of Figures \ref{fig-margin}, \ref{fig-beta}, and \ref{fig-clusterNum}, it is clearly seen that our method is considerably non-sensitive to these hyperparameters and consistently outperform the baseline of SO++ across a wide range of different values, demonstrating the robustness of our method to these hyperparameters.

\begin{table}[t]
\begin{center}
\begin{small}
\begin{tabular}{lcc}
\toprule
Network          &     Avg. of AA    & Avg. of H-score  \\
\midrule
ResNet18         &     91.3$\pm$0.2   &      87.0$\pm$0.3   \\
ResNet34         &     91.1$\pm$0.3   &      86.1$\pm$0.5   \\
ResNet50         &     91.7$\pm$0.1   &      86.8$\pm$0.2    \\
\bottomrule
\end{tabular}
\end{small}
\end{center}
\caption{Classification results on Office31 dataset (average on 6 tasks) when using different network backbones for target model, where the standard deviations are calculated by three random runs.}
\label{table-Office31-network-architature}
\end{table}

\noindent \textbf{Analysis on different target networks.} To analyse what the effects of our algorithm when using different network backbones as the target model, we experiment on Office31 dataset with other two networks of ResNet18 and ResNet34, as showed in Table \ref{table-Office31-network-architature}, where the average result and standard deviation of each task across three runs are reported. We can observe that our method is robust in using different network architectures for target model, which is very beneficial in practical applications since we can just choose a small scale network to obtain our target model of interest without loss accuracy. The small standard deviations by three runs further demonstrate robustness of our method.

\section{Discussion}
Our learning setting of UB$^2$DA is motivated by real application due to the privacy concerns and widely accessible open-AI interfaces. Our approach to learning a target model only relies on the output information from the black-box interface. As the results showed in experiments, only by accessing to the interface of source model, we can already get comparable or even better results to the previous complexity algorithms. We believe that this builds a potential guideline to the community of universal domain adaptation by rethinking whether the source data are indeed needed when transfer to a new domain. Our discovery demonstrates that a well initialized prediction model may play a more important role than source data, which inspires us when tackling a target task of interest, we should first consider to get the help from an auxiliary prediction model instead of by correcting expensive annotation data.

There are still limitations of our learning setting and algorithm. Firstly, we assume that only a single interface of source model is provided for learning, while multi-source universal black-box domain adaptation may be more practical in real application. Furthermore, the proposed learning paradigm assumes that the output information of the interfaces includes all source categorical scores, which may not hold when some of that only output the top-K prediction categories. Such problems become more complex and we leave them to future work.

\section{Conclusion}
This work provides the least restrictive setting of UB$^2$DA, where only the interface of source model is required and the target label space is allowed to be varying and unknown. To address such task, we decompose the learning objective into two subtasks of in-class discrimination and out-class detection, and then propose to unify them into a self-training framework, regularized by consistency of predictions in local neighborhoods of target samples. The proposed framework is very simple, robust, and easy to be optimized. We extensively evaluate our method on three domain adaptation benchmarks. In most of experiments, our method, learning on a black-box setting, outperforms state-of-the-art non-black-box methods with a large margin on the metric of H-score, and show comparable results to them on the metric of averaged class accuracy.

{\small
\bibliographystyle{ieee_fullname}
\bibliography{egbib}
}

\end{document}